%% file: main.tex
\documentclass{article}

\usepackage{microtype}
\usepackage{graphicx}
\usepackage{subfigure}
\usepackage{booktabs} 
\usepackage{algorithm}
\usepackage{algorithmic}
\usepackage{multirow}
\usepackage{multicol}
\usepackage{makecell}
\usepackage{bm}
\usepackage{bbding}
\usepackage{adjustbox}

\usepackage{hyperref}

\newcommand{\norm}[1]{\left\lVert #1 \right\rVert}

\input{math_commands.tex}

\usepackage[accepted]{icml2021}

\icmltitlerunning{What Do Deep Nets Learn? Class-wise Patterns Revealed in the Input Space}

\begin{document}

\twocolumn[
\icmltitle{What Do Deep Nets Learn? Class-wise Patterns Revealed in the Input Space}

\icmlsetsymbol{equal}{*}

\begin{icmlauthorlist}
Shihao Zhao$^{*}$\textsuperscript{1} \ \  
Xingjun Ma$^{* \dagger}$\textsuperscript{2} \ \ 
Yisen Wang\textsuperscript {3} \ \ 
James Bailey\textsuperscript{4} \ \ 
Bo Li\textsuperscript{5} \ \ 
Yu-Gang Jiang$^{\dagger}$\textsuperscript{1} \\
\textsuperscript{1}Fudan University \ \
\textsuperscript{2}Deakin University, Geelong \ \ 
\textsuperscript{3}Peking University \\
\textsuperscript{4}The University of Melbourne \ \
\textsuperscript{5}University of Illinois at Urbana–Champaign \\
\end{icmlauthorlist}



\icmlkeywords{Machine Learning, ICML}

\vskip 0.3in
]




\renewcommand{\thefootnote}{\fnsymbol{footnote}} 
\footnotetext[1]{Equal contribution.}
\renewcommand{\thefootnote}{\fnsymbol{footnote}} 
\footnotetext[2]{Correspondence to: Xingjun Ma (daniel.ma@deakin.edu.au), Yu-gang Jiang (ygj@fudan.edu.cn)}

\begin{abstract}
Deep neural networks (DNNs) are increasingly deployed in different applications to achieve state-of-the-art performance. However, they are often applied as a black box with limited understanding of \emph{what knowledge the model has learned from the data}.
In this paper, we focus on image classification and propose a method to visualize and understand the class-wise knowledge (patterns) learned by DNNs under three different settings including natural, backdoor and adversarial. 
Different to existing visualization methods, our method searches for a single predictive pattern in the pixel space to represent the knowledge learned by the model for each class. 
Based on the proposed method, we show that DNNs trained on natural (clean) data learn abstract shapes along with some texture, and backdoored models learn a suspicious pattern for the backdoored class.
Interestingly, the phenomenon that DNNs can learn a single predictive pattern for each class indicates that DNNs can learn a backdoor even from clean data, and the pattern itself is a backdoor trigger.
In the adversarial setting, we show that adversarially trained models tend to learn more simplified shape patterns.
Our method can serve as a useful tool to better understand the knowledge learned by DNNs on different datasets under different settings. 
\end{abstract}

\section{Introduction}
\label{Introduction}
Deep neural networks (DNNs) are a family of powerful models that have demonstrated superior learning capabilities in many real-world applications such as image classification, object detection and natural language processing.
However, after training, DNNs are often deployed as a black box with limited understanding of \emph{what knowledge the model has learned from the data}. 
Existing understandings about DNNs are mostly developed around the deep representations or the attention map. For example, DNNs are known to be able to learn high quality representations \citep{DBLP:conf/icml/DonahueJVHZTD14}, and the representations are well associated with the attention map of the model on the inputs \citep{DBLP:conf/cvpr/ZhouKLOT16,DBLP:journals/corr/SelvarajuDVCPB16}.
It has also been found that DNNs trained on high resolution images like ImageNet are biased towards texture \citep{DBLP:conf/iclr/GeirhosRMBWB19}.
While these works have significantly contributed to the understanding of DNNs, a method that can intuitively visualize the knowledge learned by DNNs for \emph{each class} is still missing.

Recently, the above understandings have been challenged by the vulnerabilities of DNNs to either backdoor \citep{DBLP:journals/corr/abs-1708-06733,chen2017targeted} or adversarial attacks \citep{szegedy2013intriguing,DBLP:journals/corr/GoodfellowSS14}.
The backdoor vulnerability is believed to be caused by the preference to learn high frequency patterns \citep{chen2017targeted,wang2020high}.
Nevertheless, no existing method has been able to reliably reveal the backdoor patterns, even though it has been well learned into the model. Adversarial attacks can easily fool state-of-the-art DNNs by either sample-wise or universal adversarial perturbations.
One recent explanation for the adversarial vulnerability is that, besides robust features, DNNs also learn useful (to the prediction) yet non-robust features that are sensitive to small perturbations \citep{DBLP:conf/nips/IlyasSTETM19}.
Adversarial training, one state-of-the-art adversarial defense method, has been shown can train DNNs to learn robust features from an image \citep{DBLP:conf/iclr/MadryMSTV18,DBLP:conf/nips/IlyasSTETM19}.
However, it is still not clear if adversarially trained DNNs can learn a single robust pattern for each class, and if so, what does the pattern look like.

In this paper, we focus on image classification DNNs and propose a visualization method that can reveal the pattern learned by the network for each class in the input (pixel) space.
Different from existing sample-wise visualization methods like attention maps, we aim to reveal the knowledge (or pattern) learned by DNNs for \emph{the entire class}. Moreover, we reveal these patterns in the \emph{pixel space} rather than the deep representation space. This is because pixel patterns are arguably much easier to interpret. Furthermore, we are interested in a visualization method that can provide new insights into the backdoor and adversarial vulnerabilities of DNNs, both of which are input space vulnerabilities \citep{szegedy2013intriguing,ma2018characterizing}.

Given a target class, a canvas image, and a subset of images from the nontarget classes, our method searches for a single pattern (a set of pixels) from the canvas image that is highly predictive of the target class. In other words, when the pattern is attached to images from any other (i.e. nontarget) classes, the model will consistently predict them as the target class.

In summary, our main contributions are:
\begin{itemize}
    \item We propose a visualization method to reveal the class-wise patterns learned by DNNs, and show its difference to universal adversarial perturbations.
    
    \item With the proposed visualization method, we show that DNNs trained on natural data can learn a consistent and predictive pattern for each class, and the pattern contains abstract shapes along with some texture.
 This sheds new light on the current texture bias question for DNNs.
 
 \item When applied on backdoored DNNs, our method can reveal the trigger pattern learned by the model from a poisoned dataset. Our method can serve as an effective tool to help identify backdoored models.
 
 \item The existence of class-wise predictive patterns indicates that even DNNs trained on clean data may have ``backdoors", and the class-wise patterns are the backdoor triggers.
 
 \item By examining the patterns learned by DNNs trained in the adversarial setting, we find that adversarially trained models tend to learn a more simplified shape pattern for each class.
\end{itemize}

\section{Related Work}
\label{Related work}

\noindent\textbf{General Understandings of DNNs.}
DNNs are known to learn more complex and higher quality representations than traditional models.
Features learned at intermediate layers of AlexNet have been found to contain both simple patterns like lines and corners and high level shapes \citep{DBLP:conf/icml/DonahueJVHZTD14}. These features have been found crucial for the superior performance of DNNs \citep{he2015delving}.
The exceptional representation learning capability of DNNs has also been found related to structures of the networks like depth and width \citep{DBLP:conf/icml/SafranS17,DBLP:conf/colt/Telgarsky16,palacio2018deep}.
One recent work found that ImageNet-trained DNNs are biased towards texture features \citep{DBLP:conf/iclr/GeirhosRMBWB19}.
Attention maps have also been used to develop better understandings of the decisions made by DNNs on a given input \citep{DBLP:journals/corr/SimonyanVZ13,DBLP:journals/corr/SpringenbergDBR14,DBLP:conf/eccv/ZeilerF14,DBLP:conf/cvpr/GanWYYH15}.
The Grad-CAM technique proposed by \cite{DBLP:journals/corr/SelvarajuDVCPB16} utilizes input gradients to produce intuitive attention maps.
Whilst these works mostly focus on sample-wise deep representations or attentions, an understanding and visualization of what knowledge DNNs learn to represent an entire class is still missing from the current literature.

\noindent\textbf{Understanding Vulnerabilities of DNNs.}
Recent works have found that DNNs are vulnerable to backdoor and adversarial attacks. 
A backdoor attack implants a backdoor trigger into a victim model by injecting (poisoning) the trigger pattern into a small proportion of training data \citep{DBLP:journals/corr/abs-1708-06733,DBLP:conf/ndss/LiuMALZW018}. The model trained on poisoned dataset will learn a noticeable correlation between the trigger pattern and a target label. A backdoored model behaves normally on clean test data, yet consistently predicts a target (incorrect) label whenever the trigger pattern appears in a test example \citep{DBLP:conf/cvpr/ZhaoMZ0CJ20,DBLP:conf/ccs/YaoLZZ19,liu2020reflection}.
This is believed to be caused by the preference to learn more high frequency (e.g. backdoor) patterns \citep{chen2017targeted,liu2020reflection,wang2020high}.
However, it is still unclear whether DNNs can learn such patterns from natural (clean) data.
Moreover, despite a few attempts \citep{DBLP:conf/sp/WangYSLVZZ19,DBLP:conf/nips/QiaoYL19}, the trigger pattern still can not be reliably revealed, even though it has been well learned into the backdoored model.

DNNs can also be easily fooled by small, imperceptible adversarial perturbations \citep{szegedy2013intriguing,DBLP:journals/corr/GoodfellowSS14,sabour2015adversarial}. Adversarial perturbations can be either sample-wise \citep{DBLP:conf/iclr/MadryMSTV18} or universal (class-wise or dataset-wise) \citep{moosavi2017universal}.
This has been found to be caused by learning useful (to prediction) but nonrobust (to small perturbations) features \citep{DBLP:conf/nips/IlyasSTETM19}. 
Meanwhile, adversarial training has been shown can train DNNs to learn more robust features from an input image~\citep{DBLP:conf/iclr/MadryMSTV18}.
However, existing understandings of adversarial training are all established around sample-wise attention maps \citep{DBLP:conf/nips/IlyasSTETM19}.
It is still not clear, from the class-wise perspective, what robust input patterns look like.
In this paper, we will propose a method to reveal the class-wise patterns learned by DNNs under both backdoor and adversarial settings.

\section{Proposed Visualization Method}
\label{Framework}
In this section, we first define the class-wise pattern searching problem, then introduce our proposed searching method.

\textbf{Motivation and Intuition.} 
We focus on image classification DNNs. We denote the training and test datasets as $\gD_{train}$ and $\gD_{test}$, respectively.
Given a DNN model $f$ trained on a $K$-class training dataset $\gD_{train}$ and a target class $y \in \{1, \cdots, K\}$, our goal is to find an \emph{input space} pattern (a small set of pixels) that are highly predictive of the target class.
A predictive pattern of a class aims to  capture the knowledge the model has learned for the class.
Intuitively, a predictive pattern of a target class should be able to make the model consistently predict the target class whenever it appears, even on other classes of images.

\noindent\textbf{Class-wise Pattern Searching.} 
Our search is based on a canvas image.
For a target class $y$, our method searches for a predictive pattern $\vp_y$ from a canvas image $\vx_c$, based on a small test subset $\gD_{n}$ of images from the nontarget classes ($\gD_{n} \subset \gD_{test}$).
The canvas image $\vx_c$ is the image where the pattern (a set of pixels) is extracted.
The search is done via an optimization process based on a mixed input between the canvas image $\vx_c$ and an image $\vx_{n} \in \gD_{n}$. The mixed input $\tilde{\vx}$ is defined as:
\begin{equation}\label{eq:1}
    \tilde{\vx}=\vm*\vx_{c}+(1-\vm)*\vx_{n},
\end{equation}
where $\vm$ is a mask that has the same size as either $\vx_c$ or $\vx_{n}$, and $\vm_{ij} \geq 0$. 
During searching, the mixed input image is labeled as the target class $y$ regardless of its original class.  This mixing strategy is reminiscent of the mixup \citep{zhang2018mixup} data augmentation.  However, we do not mix the class labels and our purpose is for pattern optimization rather than data augmentation.

During the searching process, the mask is iteratively updated to minimize the following loss:
\begin{equation}\label{eq:2}
    \gL= - \log f_y(\tilde{\vx}) + \alpha \frac{1}{n}\norm{\vm}_1,
\end{equation}
where $f_y$ denotes the network's probability output with respect to  target class $y$, $\norm{\cdot}_1$ is the $L_1$ norm, $\alpha$ is a parameter balancing the two loss terms, and $n$ is the size of the input image as well as the mask. The first loss term is the commonly used cross entropy loss. The second term increases the sparsity of the mask as we are interested in more interpretable patterns with a small number of highly predictive pixels.

During the searching process, we pair the canvas image $\vx_c$ randomly with images from $\gD_{n}$, and iteratively update the mask $\vm$ using standard Stochastic Gradient Decent (SGD) while keeping the model parameters unchanged. At each iteration, the mask $\vm$ will also be clipped into $[0, 1]$. 
Once the mask is learned, we further clip the values in the mask with respect to an additional parameter $\gamma$: smaller values than $\gamma$ are clipped to zero while larger values than $\gamma$ are clipped to one. We denote this clipped mask by $\vm_{\gamma}$.
We then extract the pattern from the canvas image by $\vp_y = \vm_{\gamma} * \vx_c$. 
The $\gamma$ parameter can be flexibly determined in different applications based on the amount of information one wants to reveal.
A large $\gamma$ will produce more sparse patterns while a small $\gamma$ will produce more fine-grained patterns.
Note that small $\gamma$ may raise the risk of the pattern overfitting to the canvas image.

The above search method is repeatedly applied to $N$ canvas images to generate $N$ patterns for each class. We then select the pattern that has the lowest loss value as the final pattern of the class. This additional step is to find the most representative (predictive) pattern by exploring more diverse canvases. 
The complete procedure of our method is described in Appendix A.

\noindent\textbf{Canvas Sampling.}
We propose four different sampling strategies for the selection of the $N$ canvas images: positive sampling, negative sampling, random sampling and white canvas.

Positive sampling selects the top-$N$ confident images from the target class according to the logits output of model $f$. Negative sampling selects the top-$N$ most confidently misclassified images from any nontarget class into the target class. The random sampling randomly chooses $N$ images from the dataset. The white canvas simply uses an image with all white pixels as the canvas. 

Both the positive and the negative sampling aim to find the most well-learned examples by the model, but from different perspectives: well-learned correctly (e.g. positive) vs. well-learned incorrectly (e.g. negative). The white canvas is interesting since the pattern found from the white canvas will have the texture (e.g. color) ``removed'', which is useful for scenarios where only the shapes are of interest. The patterns found based on different canvases are compared in Figure \ref{l_fig_5}.
After applying our method on each class, we can obtain a set of class-wise patterns: $\gP=\{\vp_1, \cdots, \vp_K\}$.
This set of predictive patterns can revel the knowledge learned by model $f$ for each class. 

\noindent\textbf{What Makes a Good Canvas?}
The canvas plays an important role in our algorithm as a constraint or regularization on the search space for finding the pattern. 
Identifying the quality and best choice for a canvas is challenging. We believe a good canvas should be a neutral or unbiased image. It can be a natural image or a white image or a random image (a space where all pixels are equally important). 
In other words, the canvas should not be a result of some optimization process, especially a process that has been targeted to fool the model.
If a canvas has been learned or is the result of adversarial perturbation, the resulting pattern will overfit to the optimization objective or be biased towards the most vulnerable (under-learned) region of the input space. 

\begin{figure*}[t]
\begin{center}
\includegraphics[width=0.95\linewidth]{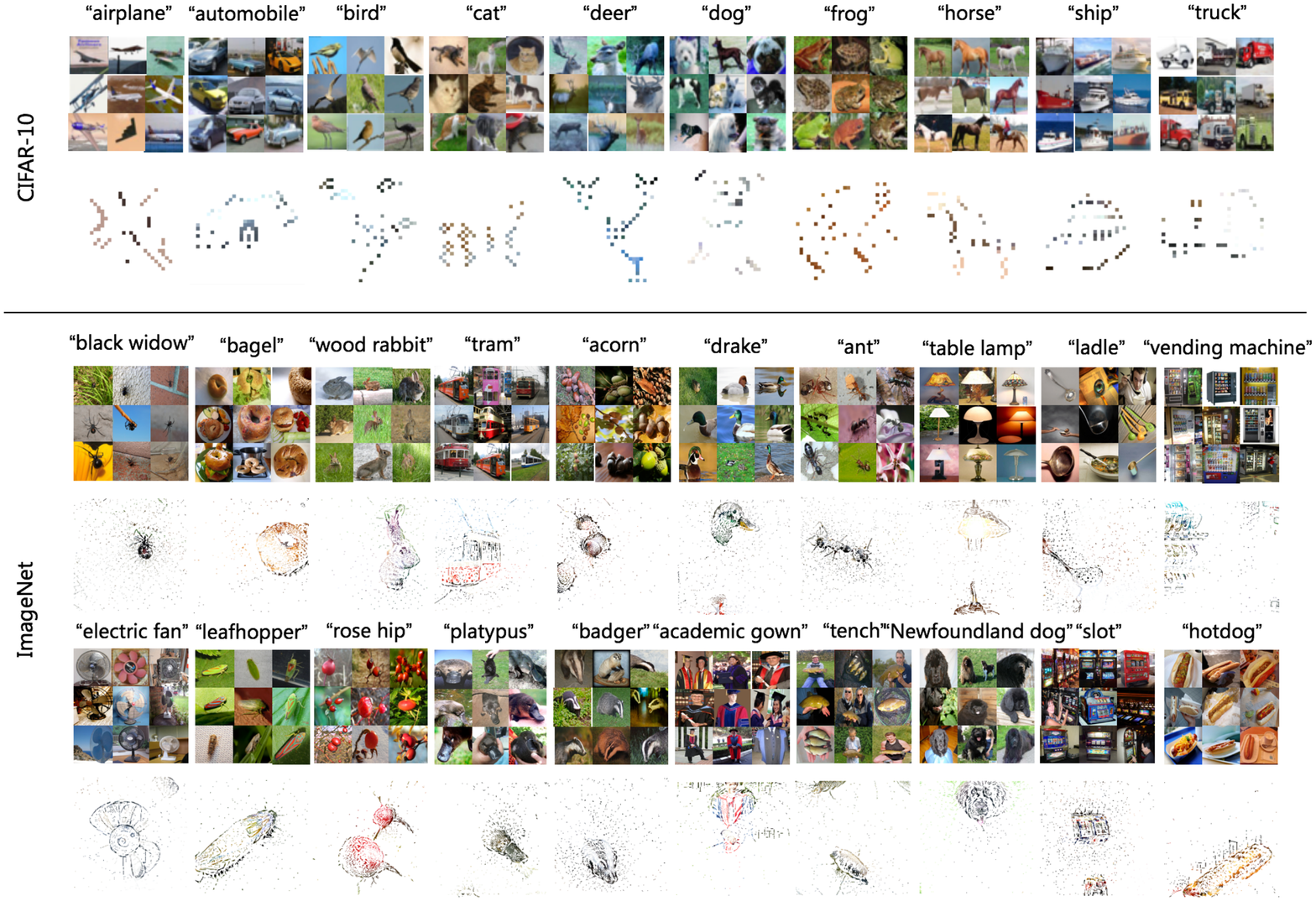}
\end{center}
   \caption{Class-wise patterns learned by naturally trained ResNet-50 models for 10 CIFAR-10 and 20 ImageNet classes. The predictive powers of these patterns are reported in Table 1 and Table 2 in Appendix. Positive canvases are used and the pattern size is 5\% image size.}
\label{l_fig_6}
\end{figure*}

\noindent\textbf{Why is it Class-wise?} At first sight, one might wonder if the discovered pattern could be sample-wise, rather than class-wise, given the use of the canvas sample. Note that, however, even though we are using a single sample as a canvas, the pattern found by the optimization algorithm is dependent on how the model has learnt the entire class, in terms of its loss. This is particularly evident in the case of the all white canvas, which bears no relation to any input sample. Hence our designation of the pattern as being ``class-wise".
While our method can find consistent and predictive class-wise patterns in the experiments, it might still be extendable.  For example, using multiple positive canvas images at the same time, using noise rather than the non-target images, or using universal adversarial perturbation (UAP) \citep{moosavi2017universal} but in an unbiased manner. 
We leave further explorations of these methods as our future work.

\noindent\textbf{Difference to Universal Adversarial Perturbation.}
UAP can also be applied to craft class-wise adversarial patterns that can make the model predict an adversarial target class.
In this view, both UAP and our method find predictive patterns to the target class.
However, the two methods work in different ways. By fooling the network, UAP explores the \emph{unlearned} (vulnerable) space (low-probability ``pockets") of the network \citep{szegedy2013intriguing,ma2018characterizing}.
In contrast, our method is a searching (rather than perturbing) method that does not rely on adversarial perturbations. Thus, it has to find the optimal pixel locations in the input space that are \emph{well-learned} by the model for the pattern to be predictive.
In Section \ref{sec:4.2} and Appendix E, we have experiments showing the difference of the patterns found by class-wise UAP and our method.

\section{Experiments}
\label{Experiments}
In this section, we will apply our method to reveal the class-wise patterns learned by DNNs under natural, backdoor and adversarial settings. We also conduct comparisons with universal adversarial perturbation and learned canvas.

\begin{figure*}[h]
\centering
\includegraphics[width=0.9\linewidth]{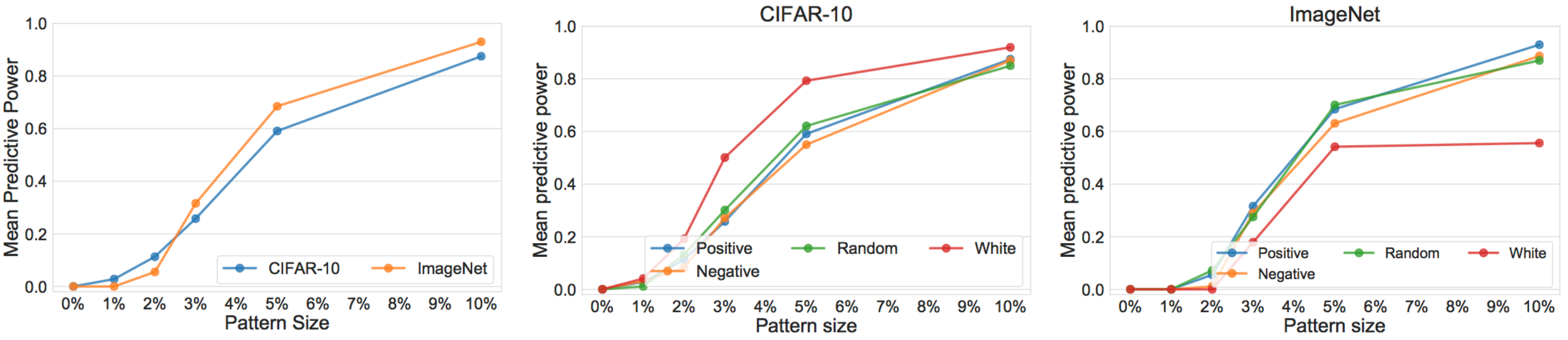}
\vspace{-0.1in}
   \caption{\textit{Left}: Mean predictive power of the class-wise patterns of different sizes, over all 10 CIFAR-10 classes and 50 randomly selected ImageNet classes. \textit{Middle}: Mean predictive power of the patterns found with different types of canvases on CIFAR-10. \textit{Right}: Mean predictive power of the patterns found with different types of canvases on ImageNet. The patterns are searched for naturally trained ResNet-50 models CIFAR-10 and ImageNet.}
\label{l_fig_3}
\vspace{-0.1in}
\end{figure*}

\begin{figure}[t]
\centering
\includegraphics[width=1.0\linewidth]{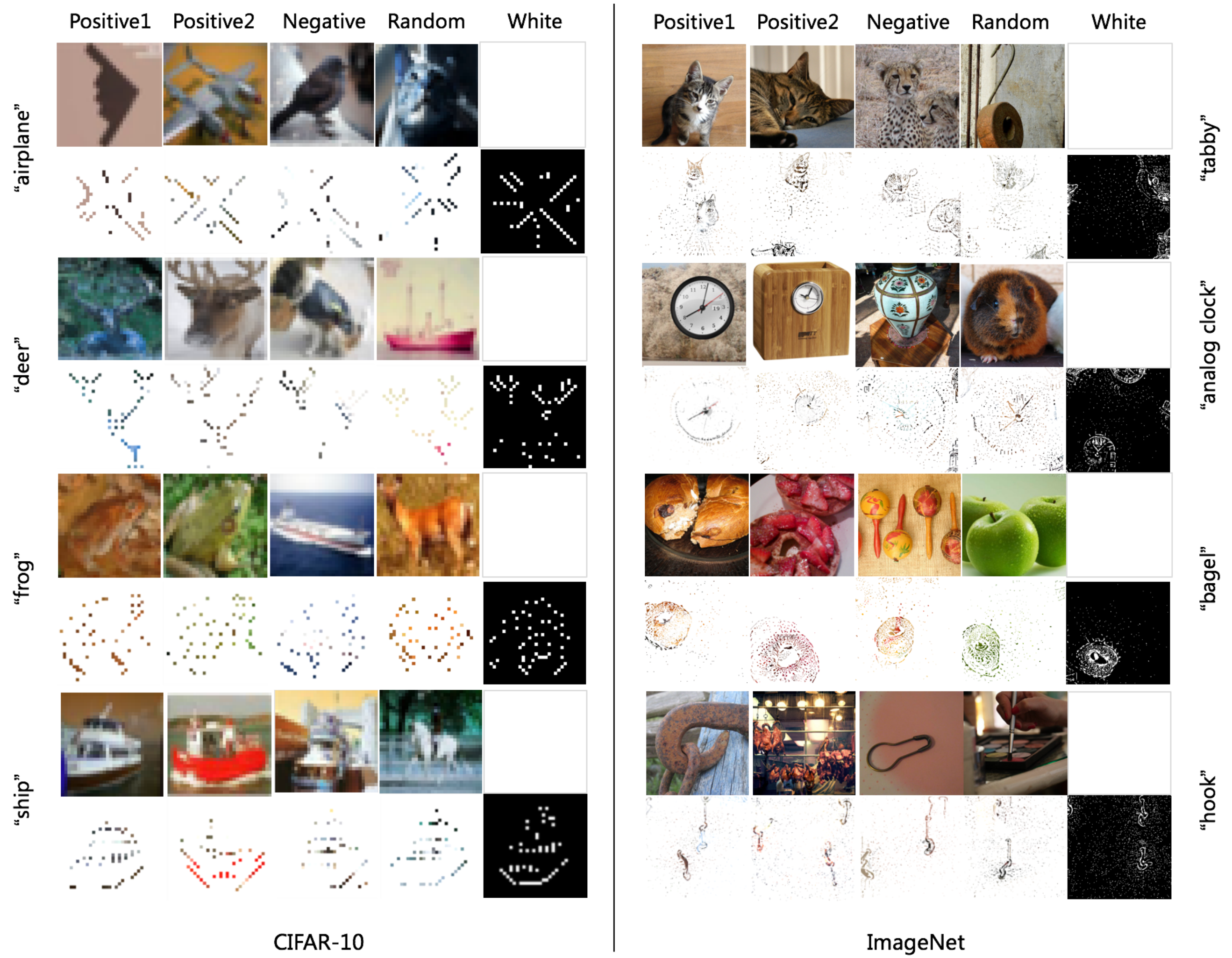}
\vspace{-0.1in}
\caption{Class-wise patterns revealed by our method using different canvases (e.g. two positive, one negative, one random and one white canvases) on CIFAR-10 and ImageNet for a naturally trained ResNet-50. The pattern size is fixed to 5\% of the image size.}
\label{l_fig_5}
\vspace{-0.1in}
\end{figure}

\noindent\textbf{Experimental Setting.}
We consider ResNet-50 \citep{DBLP:conf/cvpr/HeZRS16} on two benchmark image datasets: CIFAR-10 \citep{krizhevsky2009learning} and ImageNet \citep{imagenet_cvpr09}.
The models are trained on the training set of the datasets using standard training strategies, except the adversarial experiments in Section \ref{sec:4.4}. We then apply our method to search for the class-wise patterns based on canvas images and nontarget-class images selected from the test sets of the datasets. Here, we set the nontarget-class subset $\gD_n$ to be 20\% of the test images (more results with different sizes of $\gD_n$ can be found in Appendix C).
The sparsity regularization parameter $\alpha$ is slightly adjusted around 0.2 for different classes and datasets. The clipping parameter $\gamma$ is set to discover a pattern of a predefined size (e.g. 5\% or 10\% of the image size), which will be explicitly stated in each experiment. For the rest of this paper, we refer the pattern size by percentage, e.g., a pattern size 5\% means the pattern size is 5\% of a full image size.
The searching process is run for 5 epochs with respect to the nontarget-class subset. 

\noindent\textbf{Predictive Power.} The pattern $\vp_y$ is searched on $\gD_n \subset \gD_{test}$ (i.e. $|\gD_n|/|\gD_{test}|$=0.2) and its predictive power is tested on $\gD_{test}\setminus \gD_n$. We define the predictive power of the pattern $\vp_y$ for target class $y$ as follows: $PW = \frac{ACC(f(\vx'_n + \vp_y), y)}{ACC(f(\vx'_y), y)}$,
where $ACC(f(\vx'_n + \vp_y), y)$ is the model's accuracy on nontarget images $\vx'_n \in \gD_{test}\setminus \gD_n$ when attached with the target-class pattern $\vp_y$, and $ACC(f(\vx'_y), y)$ is the model's original accuracy on the target class images and $\vx'_y \in \gD_{test}\setminus \gD_n$. Intuitively, the $PW$ metric reflects the degree to which the pattern can represent the model's prediction performance on the true target class.

\subsection{Patterns Learned by Naturally Trained DNNs}
\label{sec:4.1}
We first show the class-wise patterns learned by ResNet-50 on natural CIFAR-10 and ImageNet datasets. Here, we use canvases sampled using the positive sampling (see Section \ref{Framework}). A more detailed analysis on the impact of different positive canvases can be found in Appendix D.
Patterns for 10 CIFAR-10 and 20 (out of the 50 randomly selected) ImageNet classes at 5\% image size are provided in Figure \ref{l_fig_6}. The predictive power of different pattern sizes (0\% - 10\%) are shown in Figure \ref{l_fig_3}. The detailed predictive power results for each of the 10 CIFAR-10 classes and 50 randomly selected ImageNet classes can be found in Appendix B.

\textbf{Patterns Revealed by Positive Canvas.}
As shown in Figure \ref{l_fig_6}, the class-wise patterns learned from natural data contain shapes that are closely related to the object class, though some of the shapes are quite abstract (e.g. the CIFAR-10 classes).
One interesting observation is that DNN can learn more than one object for a given class especially on high resolution dataset ImageNet, for example the ``acorn", ``ant", ``rose hip" and ``slot" classes. More interestingly, many of the objects learned by the network are not located at the center of the input space.
We conjecture this is because the same class of objects may appear at different locations of the training images.
Consequently, the model learns to remember more than one objects at various locations of the input space.
According to the mean (over all tested classes) predictive power shown in Figure \ref{l_fig_3} (left subfigure), the patterns discovered by our method at pattern size 5\% can roughly represent 60\% of the predictive information learned by the model. 
This confirms that DNNs can indeed remember some common objects for each class, even though the training images of the class are very diverse.

The learned objects have shapes and some textures (e.g. color).
Shapes are more class-wise features as our method iteratively searches for the optimal positions of the pixels that are predictive to the entire class.
By contrast, the textures are more sample-wise features carried over from the canvas image by the $\vp_y = \vm_{\gamma} * \vx_c$ operation.
The use of a white-canvas can remove these textures, but also reveals slightly different shapes (see Figure \ref{l_fig_5}). The lack of fine-grained textures in the revealed patterns indicates that DNNs tend to remember more about the shapes at different locations of the input space.
In relation to existing texture bias understanding of DNNs trained on ImageNet \citep{DBLP:conf/iclr/GeirhosRMBWB19}, our result implies that shape also plays an important role in DNNs.
Note that the $L1$ regularization used in our method only encourages the pattern to be sparse but does not restrict the pattern to be textures or shapes (textures can also be sparse).

\noindent\textbf{Patterns Revealed by Different Types of Canvases.}
A few examples of the patterns found on 4 different types of canvases are illustrated in Figure \ref{l_fig_5}. 
As can be observed, the patterns discovered on different canvases are quite similar on CIFAR-10, but it is not the case for ImageNet.
This indicates that the network learns to remember more than one patterns, especially on diverse datasets like ImageNet. And they are all highly predictive of the class (see middle and right subfigures of Figure \ref{l_fig_3}).
In some of the examples, the pattern exhibits a noticeable correlation with the canvas image, except when the white-canvases are used. 
The patterns revealed on white-canvases also contain visible shapes.
In terms of the predictive power, although the mean predictive powers are similar for different types of canvases (middle and right subfigures of Figure \ref{l_fig_3}), the individual predictive power can vary when different canvas images are used (Table 3 and Table 4, Appendix D). This is also why we suggest to use $N$ (i.e. $N$=5) different canvases to help find the most predictive patterns.

\begin{figure}[t]
\begin{center}
\includegraphics[width=0.9\linewidth]{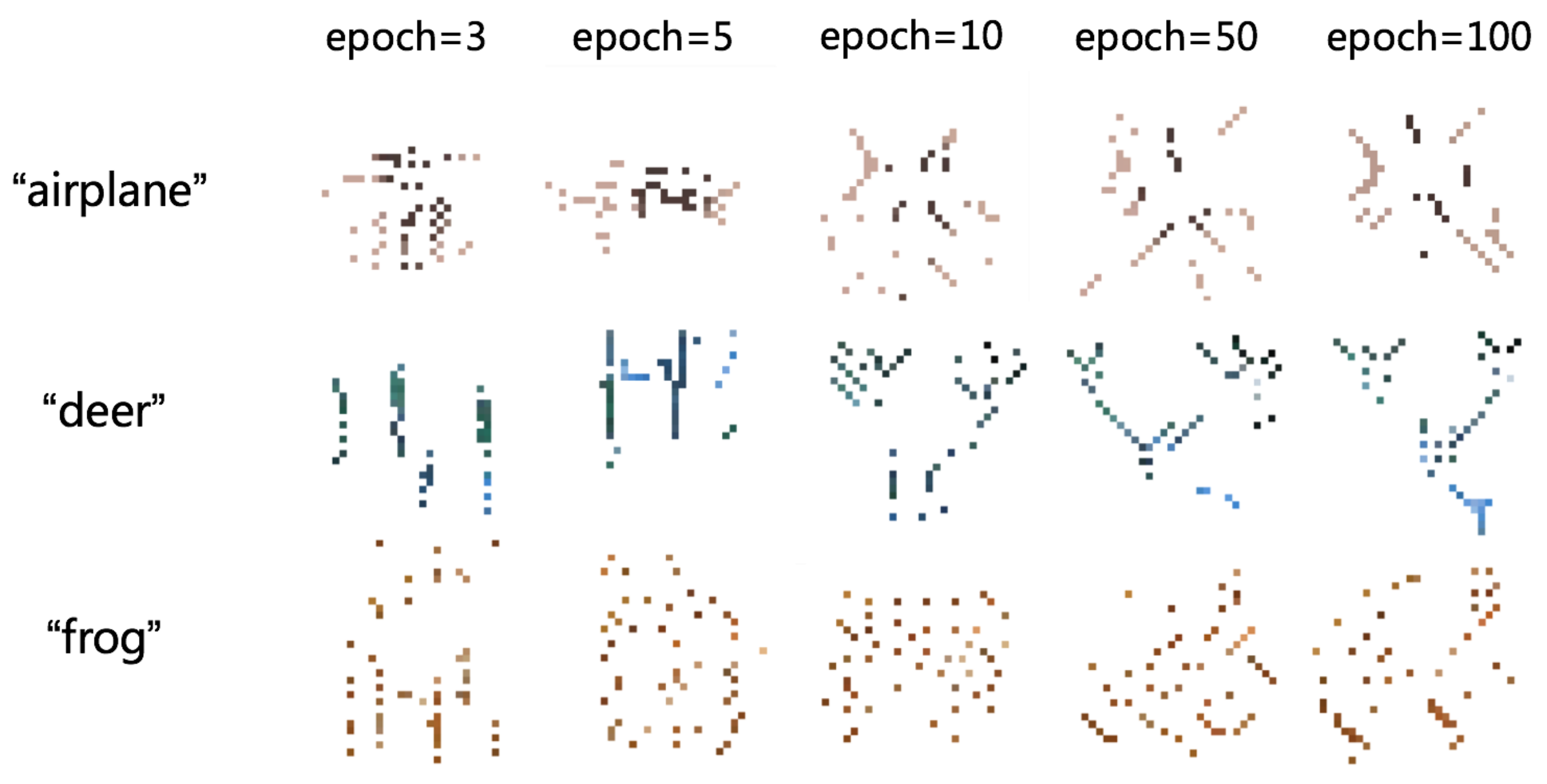}
\end{center}
   \caption{Class-wise patterns learned by a ResNet-50 at training epochs 3, 5, 10, 50 and 100 on CIFAR-10 dataset and the corresponding test accuracies of the model are 31.46\%, 52.43\%, 69.26\%, 79.12\% and 94.24\%, respectively.}
\label{l_fig_14}
\end{figure}

\noindent\textbf{Patterns Learned at Different Training Stages.}
\label{sec:appd_h}
It is interesting to visualize the different levels of patterns learned by DNNs at different training stages. Here, we apply our method to a ResNet-50 trained on CIFAR-10 and visualize the patterns learned at different epochs in Figure \ref{l_fig_14}.
Interestingly, the network learns some basic horizontal or vertical patterns at epoch 3, then connections between the basic patterns at epoch 5, some simple object shapes at epoch 10, and fine-grained shapes at the later training stage (epoch 50/100). The different levels of patterns correspond well to the model's test accuracy. We believe these patterns can help better understand the learning dynamics of DNNs.

\subsection{Comparison to UAP and Exploration of Alternative Searching Methods}
\label{sec:4.2}
Here, we compare the patterns revealed by Universal Adversarial Perturbation (UAP) and our method. We also explore alternative searching strategies with 1) our method plus the total-variation (TV) regularization \citep{fong2017interpretable} and 2) our method with learned canvas. These comparisons and explorations can help better understand our proposed method and different alternatives.

\noindent\textbf{Comparison to UAP.}
We first compare our method to UAP and the TV variant of our method.
We apply the UAP targeted adversarial perturbation \cite{moosavi2017universal} separately for each class. The class-wise UAP is crafted on the entire test set of CIFAR-10 and 10K randomly selected test images for ImageNet. The TV variant of our method utilizes the loss function:
\begin{equation}\label{eq:3}
    \gL= - \log f_y(\tilde{\vx}) + \alpha \frac{1}{n}\norm{\vm}_1 + \beta \frac{1}{n}\norm{\nabla \vm}^2_2,
\end{equation}
where $\beta$ is a parameter balancing the TV regularization (the third term).
The reason why we consider the TV regularization is that it can reduce the variation of the mask, producing more smooth patterns \citep{fong2017interpretable}. 

\begin{figure}[t]
\centering
\includegraphics[width=0.95\linewidth]{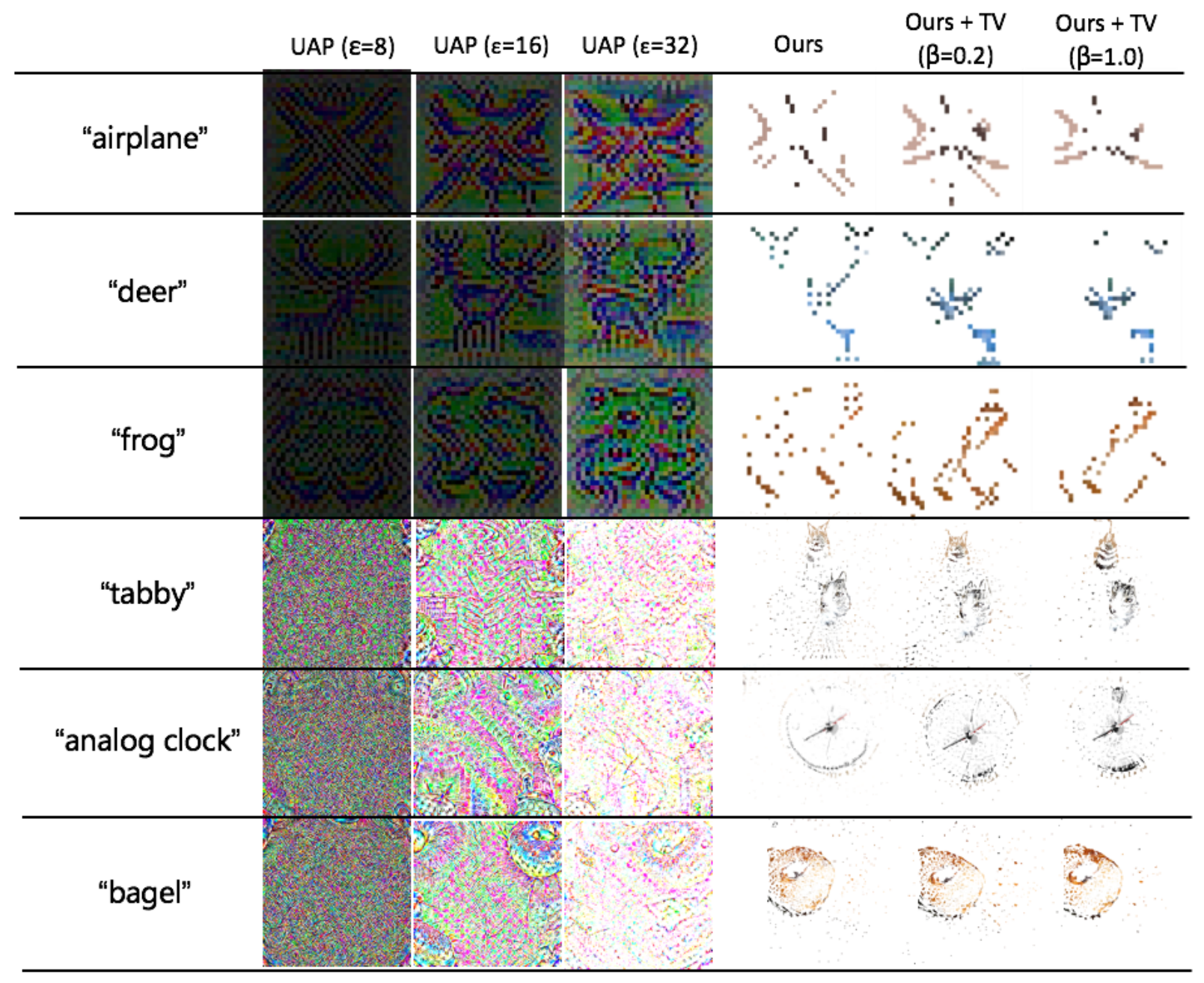}
\vspace{-0.1in}
   \caption{A comparison of the class-wise patterns revealed on naturally trained ResNet-50 models by universal adversarial perturbation (UAP), our method and our method with the total-variation (TV) regularization (defined in \Eqref{eq:3}). For our methods, the pattern size is set to 5\% of the image size, while UAP perturbs the entire (i.e. 100\%) image. The top and bottom three rows show the patterns for three CIFAR-10 and ImageNet classes, respectively.}
\label{l_fig_4}
\vspace{-0.1in}
\end{figure}

As shown in Figure \ref{l_fig_4}, the UAP patterns generated on CIFAR-10 also contain structures of the object. However, the patterns found by our method are much cleaner. The UAP patterns generated on ImageNet are notably more abstract than our methods. This is because, by fooling the network, 
UAP explores the \emph{unlearned} (vulnerable) space of DNNs. 
By contrast, our method searches for the \emph{well-learned} locations in the input space to find the most predictive patterns to the class.
We believe UAP should be applied in an unbiased manner without interfering with the exploration of the learned space of the network. We find that the TV regularization can indeed help produce smoother patterns, however, the main patterns are generally similar to those found without TV regularization. The predictive powers of these methods are analyzed in Appendix E.

\noindent\textbf{Exploring Learned Canvases.}
\label{sec:appd_i}
Here, we adapt our method to simultaneously perturb the mask $\vm$ and the canvas $\vx_c$ as follows:
\begin{equation}\label{eq:4}
    \min \limits_{\vx_{c}, \vm} -\log f_y(\vm*\vx_{c}+(1-\vm)*\vx_{n}) + \alpha \frac{1}{n}\norm{\vm}_1.
\end{equation}
This is a straightforward extension of our method defined in \Eqref{eq:2}, with the canvas $\vx_c$ is also perturbed during the searching process. This extension can also be interpreted as a combination of our searching algorithm with UAP applied on the canvas image $\vx_c$. The above objective is solved by an alternating optimization strategy: the mask and the canvas are optimized for 5 steps alternatively with step size $2/255$ and 0.02 respectively. The same normalization and clipping techniques are used here as in our original method. We use positive sampling to select the initial canvas image and set the pattern size to 5\%.

\begin{figure}[t]
\begin{center}
\includegraphics[width=0.9\linewidth]{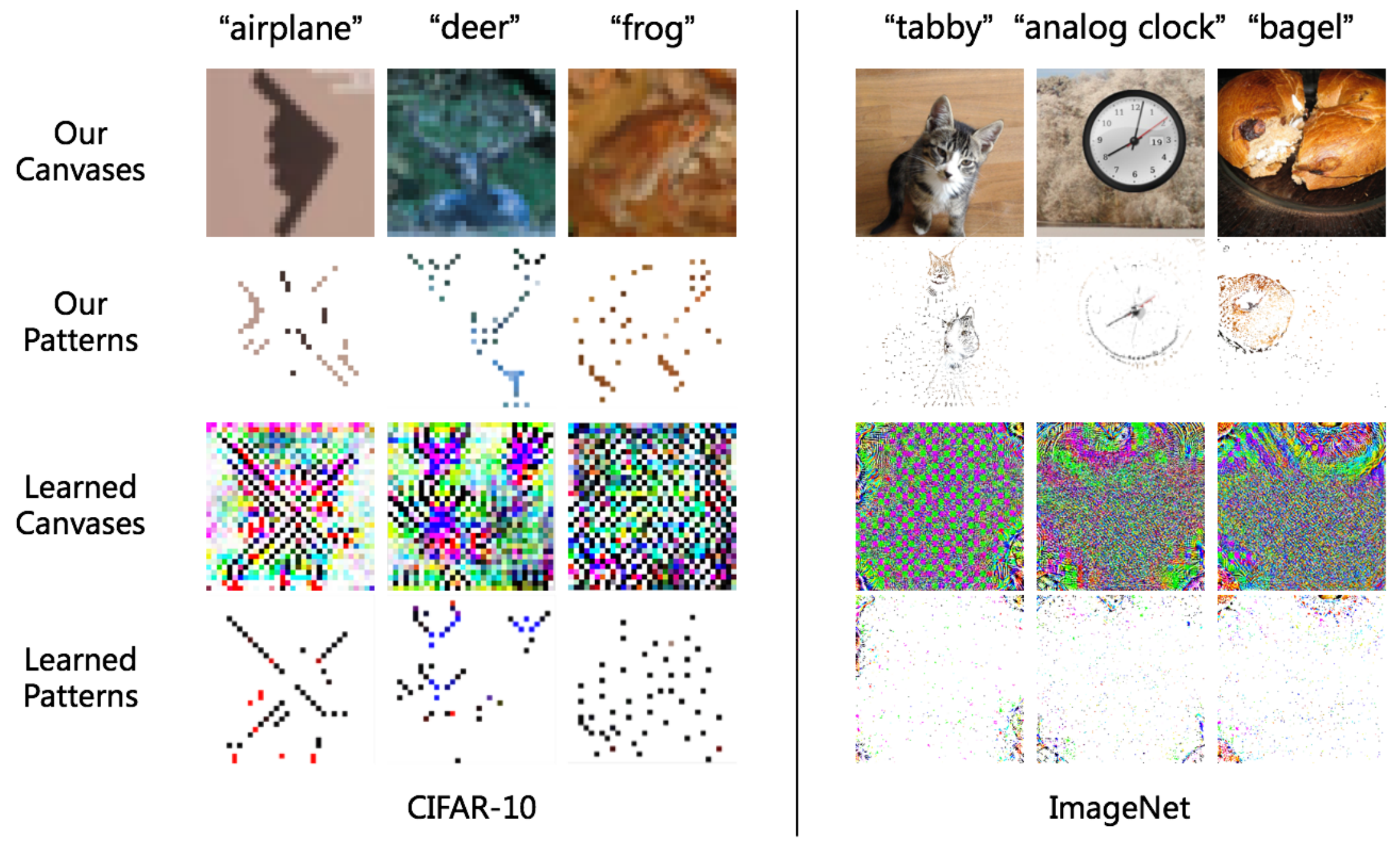}
\end{center}
   \caption{A comparison of the class-wise patterns revealed by our method and an learned canvas version of our method (defined in \Eqref{eq:4}) for naturally trained ResNet-50s on CIFAR-10 and ImageNet. The pattern size is fixed to 5\% of the image size. The first row shows the patterns found by the original method, while the learned canvas and the revealed patterns are shown in the second and third rows respectively.}
\label{l_fig_12}
\end{figure}

We compare side-by-side the patterns revealed by our original method and the learned canvas variant in Figure \ref{l_fig_12}. 
On CIFAR-10 dataset, the patterns revealed on learned canvas also contain object shapes, but for some classes, it only reveals part of the object. For example, the antlers for the ``deer" class. This phenomenon is more obvious on ImageNet, where
it fails to reveal any meaningful patterns. 
Notably, the pixels found on the learned canvas  are mostly located at the four corners of the input space (the right panel of Figure \ref{l_fig_12}).
This is because those regions are the most under-learned and vulnerable regions where the loss can be effectively minimized towards the target class. This indicates that learned canvases can be easily biased to explore the unlearned space of DNNs, which goes against the purpose of the visualization -- revealing the well-learned space by the network. The corresponding predictive power results are reported in Table 6 (Appendix E). The conclusion is that learned canvases do not necessarily lead to more predictive patterns. It is still an open question as for how a canvas can be learned without causing any biases.

A comparison of our method to the sample-wise attention maps found by Grad-CAM \citep{DBLP:journals/corr/SelvarajuDVCPB16} has also been provided in Appendix F.
These explorations confirm that our proposed method in \Eqref{eq:2} is a reasonable method for revealing the patterns learned by DNNs.

\subsection{Patterns Learned by Backdoored DNNs}\label{sec:4.3}
\begin{figure}[t]
\centering
\includegraphics[width=0.95\linewidth]{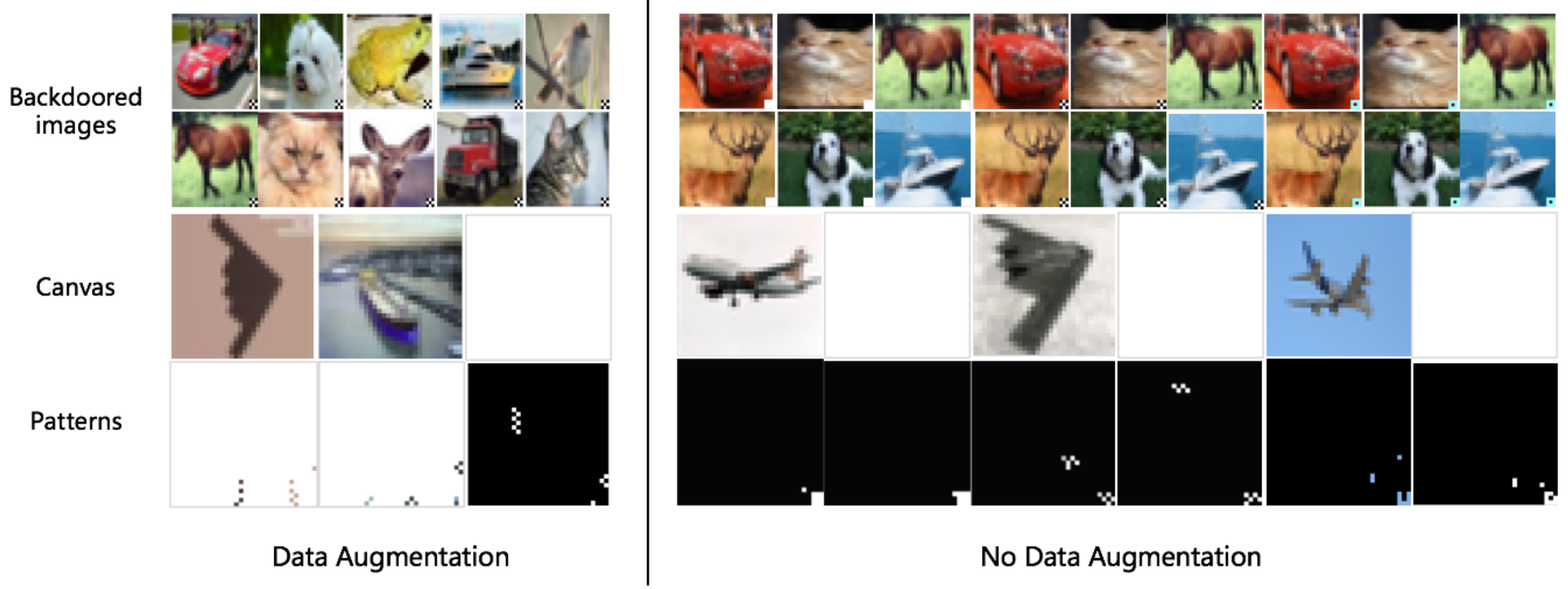}
\vspace{-0.1in}
   \caption{Backdoor patterns revealed by our method for different backdoor triggers. The backdoored ResNet-50 is trained on CIFAR-10 and poisoned by BadNets with target class ``airplane". The pattern size is 1\%. Data augmentation means the backdoored model was trained with standard data augmentation techniques.}
\label{l_fig_7}
\vspace{-0.1in}
\end{figure}

Here, we apply our method to reveal the patterns learned by a backdoored ResNet-50 model by BadNets \citep{DBLP:journals/corr/abs-1708-06733} on CIFAR-10. We test two models that trained with and without data augmentations including random shift, random crop and random horizontal flip.
The patterns revealed on different canvases are illustrated in Figure \ref{l_fig_7}. Here, we set the pattern size to be small (i.e. 1\%) as we are interested in examining whether the model has learned a small but extremely predictive pattern (e.g. the trigger pattern).
As can be observed, our method can reliably reveal the trigger pattern for backdoored models trained either with or without data augmentations, although the locations are slightly different to the original trigger locations. 
In the no data augmentation case, the recovered triggers are very close to the ground truth. This indicates that DNNs can memorize both the trigger pattern and its location.
An interesting observation is that, in some of the recovered patterns, there are also similar trigger patterns that appear at various locations of the input space. For example, the middle two columns in the right panel of Figure \ref{l_fig_7}. This indicates that the model may learn similar trigger patterns from \emph{natural} data.
As shown in the left panel of Figure \ref{l_fig_7}, data augmentations tend to shift both the shape and the location of the trigger learned by the model, as expected.

\begin{figure}[t]
\centering
\includegraphics[width=0.95\linewidth]{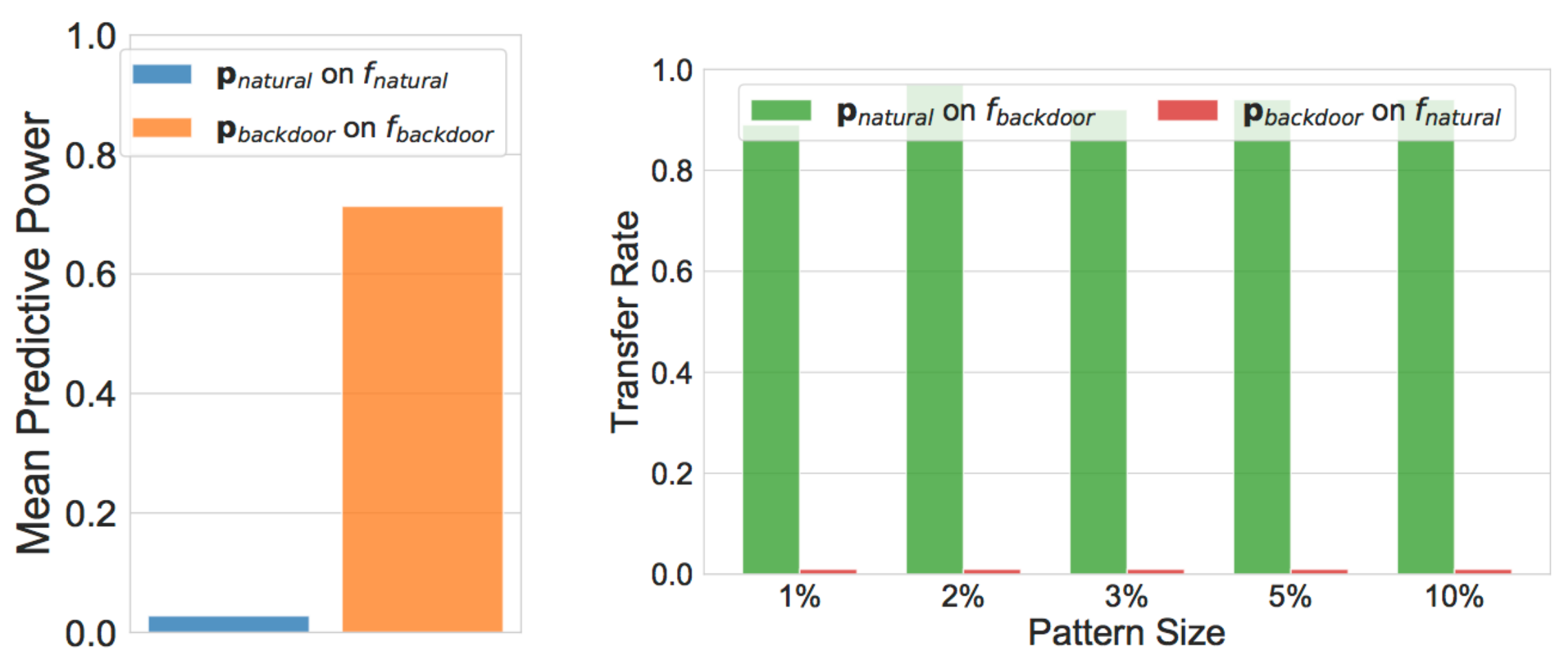}
\vspace{-0.1in}
   \caption{
   \emph{Left}: The mean (over all 10 CIFAR-10 backdoor target classes) predictive power of 1) the recovered trigger pattern (1\% image size) on the backdoored model ($\vp_{backdoor}$ on $f_{backdoor}$), and 2) the same size (1\% image size) of natural pattern on natural model ($\vp_{natural}$ on $f_{natural}$).  \emph{Right}: The transferability of the patterns revealed for clean ResNet-50 models. The transfer rate is defined as $PW_{source}/PW_{target}$, the ratio between the predictive power on the source and the target model. The patterns are searched on the source model.}
\label{l_fig_8}
\vspace{-0.1in}
\end{figure}

We then test the predictive power of the recovered triggers on the backdoored model ($f_{backdoor}$). For a comparison, we also test the predictive power of the same size of pattern for a naturally-trained model ($f_{natural}$). Here, we apply BadNets to attack all CIFAR-10 classes individually, and compute the mean predictive power over all classes (the triggers are still recovered separately for each class). As shown in the left bar plot of Figure \ref{l_fig_8}, the patterns are highly predictive on the backdoored model, yet the same size of patterns learned by the naturally-trained ResNet-50 has almost zero predictive power. We also cross test various sizes of the patterns learned by $f_{backdoor}$ and $f_{natural}$ on each other, and show the transfer rate of the predictive power in the right bar plot of Figure \ref{l_fig_8}.
We find that the patterns learned by $f_{natural}$ can always transfer to $f_{backdoor}$, however, the opposite direction does not work.
The existence of small, highly predictive but  non-transferable patterns indicates that the model has learned something unnatural (e.g. a backdoor trigger).
The effectiveness of our method on more complex backdoor triggers is worth further exploration.

Revisiting the patterns learned from natural data in Figure \ref{l_fig_6}, we find that DNNs trained on natural data may also have ``backdoors", since those class-wise patterns can be immediately applied to ``backdoor" attack the model.
According to the predictive power results in Figure \ref{l_fig_3}, the attack success rate will be high, although not as high as the state-of-the-art \citep{chen2017targeted,liu2020reflection}.
Our finding that DNNs can memorize a single predictive pattern for the entire class explains why backdoor attacks can easily succeed by poisoning only a small proportion of the training data.

\subsection{Patterns Learned by Adversarially Trained DNNs}
\label{sec:4.4}
Here, we apply our method to reveal the class-wise patterns learned by ResNet-50 on three different versions of CIFAR-10 dataset: natural ($\gD$), robust ($\gD_{R}$) and nonrobust ($\gD_{NR}$).
The $\gD_{R}$ and $\gD_{NR}$ versions were generated in previous work \citep{DBLP:conf/nips/IlyasSTETM19} by perturbing the training images of the original dataset $\gD$ to have only robust or nonrobust features, under an adversarial setting. 
In other words, $\gD_{R}$ contains only the robust features learned by a robust (adversarially trained) model, while $\gD_{NR}$ contains only the nonrobust features learned by a nonrobust (naturally trained) model.
It has been shown that DNNs exhibit moderate adversarial robustness when trained on $\gD_{R}$ even using natural training \citep{DBLP:conf/nips/IlyasSTETM19}. 

\begin{figure}[t]
\centering
\includegraphics[width=0.95\linewidth]{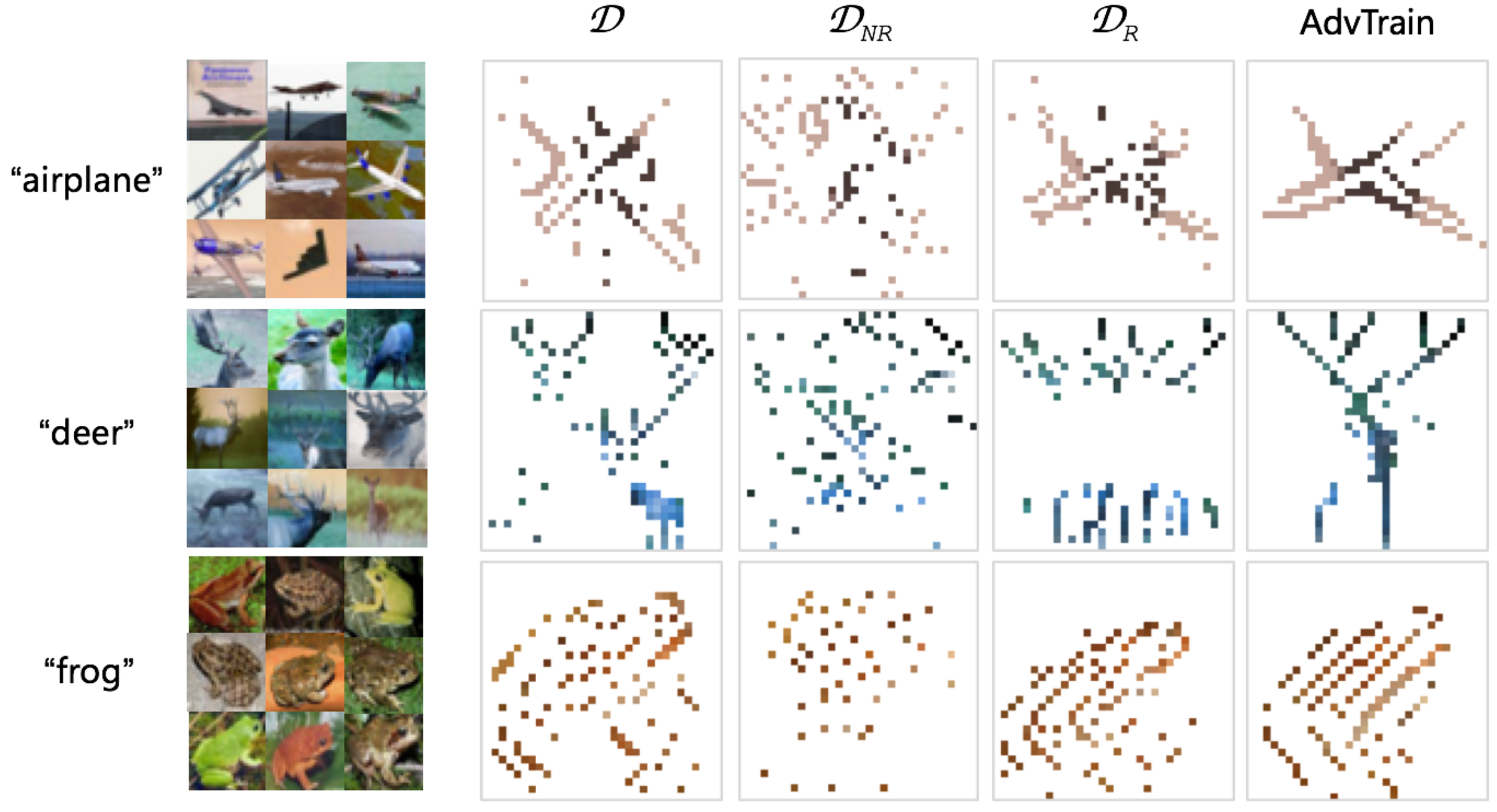}
\vspace{-0.1in}
   \caption{Examples of class-wise patterns learned by ResNet-50 and revealed by our method on natural ($\gD$), nonrobust ($\gD_{NR}$) and robust ($\gD_{R}$) data on CIFAR-10, as well as adversarially trained model (AdvTrain) on CIFAR-10. Pattern size is 10\% image size.
   }
\label{l_fig_9}
\vspace{-0.1in}
\end{figure}

We train the ResNet-50 model independently using natural training on $\gD$, $\gD_{NR}$ and $\gD_{R}$, and the adversarial training on $\gD$ following \cite{DBLP:conf/iclr/MadryMSTV18}.
The class-wise patterns learned by these models are illustrated in Figure \ref{l_fig_9}. 
Compared to $\gD$, the patterns learned on $\gD_{NR}$ are mostly background noise.
However, patterns learned by the adversarially trained model contains clear shapes and much less background noise. Compared to that learned on $\gD$, the shapes learned by adversarially trained DNNs are more simplified and have more densely connected regions.
This provides an intuitive visualization of the robust patterns learned by adversarially trained DNNs from a class-wise perspective.
The robust dataset $\gD_{R}$ is interesting as it is a dataset that is intentionally perturbed to be robust. 
The patterns learned on $\gD_{R}$ reveal that the ``robust" perturbation can indeed remove background noise (e.g. the ``airplane" class).  However, it may also remove a certain part of the object shape (e.g. the entire body of the ``deer"). 
This might produce DNNs that have poor generalization performance in real-world scenarios.

\section{Discussion and Conclusion}
\label{Conclusion}
In this paper, we propose a method to reveal the \emph{class-wise} patterns learned by deep neural networks (DNNs) in the \emph{input space}.
The patterns revealed by our method indicate that DNNs can learn to memorize object shapes. 
By further examining the patterns learned by backdoored DNNs, we find that our method can reveal the trigger pattern, and more interestingly, DNNs trained on clean data may also have ``backdoors".
Patterns revealed in adversarial settings indicate that adversarially trained DNNs can indeed learn more robust shape patterns, however, internationally-perturbed robust datasets may lose certain shape features.
Our method can serve as a useful tool for communities to better understand DNNs trained on different datasets under different settings.

One application of our method is DNN understanding and interpretation. Based on our method, one can develop metrics to measure the strength, weakness or biases (e.g. gender, color and other attributes) of the knowledge learned by DNNs. The other potential application is using our method to develop effective backdoor defense methods by monitoring and avoiding the learning of highly predictive patterns during training. Our method can also motivate more effective adversarial defense methods, for example, regularized adversarial training methods that can help DNNs learn more robust shape features.

\bibliography{icml2021}
\bibliographystyle{icml2021}

\clearpage
\appendix

\section{Algorithm and Search Settings}
\label{sec:appd_a}

The proposed Input Space Class-wise Pattern (ISCP) searching method is described in Algorithm \ref{alg:1}. 
Given a DNN model $f$ (trained on $\gD_{train}$), a target class $y$ and a canvas image $\vx_c$ (in the case of positive and negative canvas, $\vx_c$ is sampled from $\gD_{test}$), the pattern search produce is executed as follows.
The mask $\vm$ is initialize to have the same size with the input image (without the channel dimension) and all 0.5 values (line 1). We randomly sample a subset of non-target class images $\gD_{n}$ from the test set $\gD_{test}$, which in most of our experiments, is set to the non-target class images from 20\% of the test set.

The searching procedure iterates for $T=5$ epochs ($T$ times full pass on $\gD_{n}$) with minibatch size $B=4$.
In each epoch, for each minibatch $\{\vx_n^{(j)}\}_{j=1}^{B}$ in $\gD_{n}$, we first mix each image in the minibatch with the canvas image $\vx_c$ using the mask $\vm$ (line 4). We then compute the loss on the mixed inputs (line 5), and backpropagate to update the mask according to a step size $\eta=0.02$ (lines 6-7). The updated mask is then clipped into the value range of $[0, 1]$. 

After $T$ epochs of optimization, we obtain a mask $\vm$. We then clip the values according to a threshold $\gamma$, so that only a certain proportion (e.g. 5\%) of the values are clipped to 1 (the rest are clipped to 0).
We then apply the clipped mask on the canvas image $\vx_c$ to obtain the final pattern $\vp_y$ for target class $y$.
The loss term balancing parameter $\alpha \in [0.1, 0.3]$ is slightly adjusted around 0.2 to produce the most predictive pattern. The $\gamma$ parameter is set to the value that only a certain percentage of the pixels are remained, which in most of our experiments is 5\% of the image size.

\begin{algorithm}
\caption{Input Space Class-wise Pattern (ISCP) Searching Method}
\label{alg:1}
\begin{algorithmic}[1]
\REQUIRE ~~\\
A DNN model $f$, canvas image $\vx_c$, nontarget-classes subset $\gD_{n} \subset \gD_{test}$, target class $y$, regularization balancing parameter $\alpha$, clipping threshold $\gamma$, input image size $W\times H$, epochs $T=5$, batch size $B=4$, step size $\eta=0.02$.
\ENSURE Pattern $\vp_y$ for target class $y$
\STATE $\vm$ = InitializeMask()
\FOR {$t$ in $\mathrm{range}(T)$}
\FOR {$\{\vx_n^{(j)}\}_{j=1}^{B}$ in $\gD_{n}$}
\STATE $\tilde{\vx}^{(j)}=\vm*\vx_c+(1-\vm)*\vx_n^{(j)}$
\STATE $\gL=\sum_{j=1}^{B}\{- \log f_y(\tilde{\vx}^{(j)}) + \alpha \frac{1}{n}\norm{\vm}_1\}$
\STATE $\bm{\delta}=\frac{\partial{\gL}}{\partial{\vm}}$ 
\STATE $\vm = \vm-\eta*\mathrm{sign}(\bm{\delta})$
\STATE Clip all the values in $\vm$ to [0, 1]
\ENDFOR
\ENDFOR
\STATE Clip the values in $\vm$ that are smaller than $\gamma$ to 0, larger than $\gamma$ to 1, and get $\vm_{\gamma}$ 
\STATE $\vp_y = \vm_{\gamma} * \vx_c$
\STATE return $\vp_y$
\end{algorithmic}
\end{algorithm}{}

\begin{table*}[t]
\caption{Detailed predictive power of different sizes of the patterns found by our method for ResNet-50 trained on natural CIFAR-10. The predictive power is shown separately for each CIFAR-10 class. \textbf{mPW}: mean predictive over all classes; \textbf{STD}: standard deviation over all classes. Positive canvases are used.}
\label{tab:a_1}
\begin{center}
\setlength{\tabcolsep}{4mm}{
\begin{tabular}{c|ccccc|c}
\toprule
\multirow{2}{*}{\textbf{Class}} & \multicolumn{5}{c|}{\textbf{Pattern Size}} & \multirow{2}{*}{\textbf{Original Acc}} \\
 &  \textbf{1\%} & \textbf{2\%} & \textbf{3\%} & \textbf{5\%} & \textbf{10\%} & \\
\midrule
airplane & 0.031 & 0.126 & 0.357 & 0.745 & 0.892 & 0.953\\
automobile & 0.032 & 0.081 & 0.246 & 0.663 & 0.921 & 0.979\\
bird & 0.043 & 0.161 & 0.333 & 0.665 & 0.955 & 0.932\\
cat & 0.047 & 0.077 & 0.162 & 0.431 & 0.814 & 0.851\\
dear & 0.021 & 0.104 & 0.365 & 0.698 & 0.771 & 0.960\\
dog & 0.011 & 0.099 & 0.270 & 0.571 & 0.867 & 0.910\\
frog & 0.010 & 0.041 & 0.104 & 0.383 & 0.838 & 0.967\\
horse & 0.040 & 0.151 & 0.278 & 0.596 & 0.865 & 0.948\\
ship & 0.010 & 0.124 & 0.248 & 0.589 & 0.901 & 0.968\\
truck & 0.032 & 0.167 & 0.212 & 0.570 & 0.928 & 0.957\\
\hline
\textbf{mPW} & 0.028 & 0.113 & 0.258 & 0.591 & 0.875 & 0.942\\
\textbf{STD} & 0.013 & 0.038 & 0.079 & 0.107 & 0.053 & 0.036\\
\bottomrule
\end{tabular}
}
\end{center}
\end{table*}

\begin{table*}[t]
\caption{Detailed predictive power of different sizes of the patterns found by our method for ResNet-50 trained on natural ImageNet. The predictive power is shown separately for each of the 50 ImageNet classes. \textbf{mPW}: mean predictive over all 50 classes; \textbf{STD}: standard deviation over all 50 classes. Positive canvases are used.}
\label{tab:a_2}
\begin{center}
\setlength{\tabcolsep}{3mm}{
\begin{tabular}{c|ccccc|c}
\toprule
\multirow{2}{*}{\textbf{Class}} & \multicolumn{5}{c|}{\textbf{Pattern Size}} & \multirow{2}{*}{\textbf{Original Acc}} \\
 &  \textbf{1\%} & \textbf{2\%} & \textbf{3\%} & \textbf{5\%} & \textbf{10\%} & \\
\midrule
tench & 0.000 & 0.054 & 0.293 & 0.609 & 0.641 & 0.920\\
black widow & 0.000 & 0.116 & 0.661 & 0.768 & 0.779 & 0.860\\
drake & 0.000 & 0.083 & 0.523 & 0.821 & 0.952 & 0.840\\
platypus & 0.000 & 0.000 & 0.013 & 0.275 & 0.750 & 0.800\\
dowitcher & 0.000 & 0.062 & 0.300 & 0.656 & 0.856 & 0.900\\
oystercatcher & 0.000 & 0.007 & 0.177 & 0.469 & 0.677 & 0.960\\
chesapeake bay retriever & 0.000 & 0.057 & 0.193 & 0.636 & 0.852 & 0.880\\
schipperke & 0.000 & 0.036 & 0.155 & 0.667 & 0.929 & 0.840\\
newfoundland dog& 0.000 & 0.183 & 0.695 & 0.878 & 0.902 & 0.820\\
toy poodle & 0.000 & 0.146 & 0.667 & 1.270 & 1.292 & 0.480\\
leopard & 0.000 & 0.091 & 0.625 & 0.705 & 0.659 & 0.880\\
ant & 0.000 & 0.105 & 0.776 & 0.921 & 0.987 & 0.760\\
leafhopper & 0.000 & 0.044 & 0.378 & 0.656 & 0.778 & 0.900\\
wood rabbit & 0.000 & 0.000 & 0.070 & 0.465 & 0.802 & 0.860\\
badger & 0.000 & 0.000 & 0.081 & 0.453 & 0.674 & 0.860\\
gorilla & 0.000 & 0.013 & 0.113 & 0.563 & 0.675 & 0.800\\
academic gown & 0.000 & 0.000 & 0.026 & 0.632 & 1.289 & 0.380\\
backpack & 0.000 & 0.000 & 0.045 & 0.955 & 1.841 & 0.440\\
bicycle-built-for-two & 0.000 & 0.081 & 0.302 & 0.698 & 0.814 & 0.860\\
bookcase & 0.000 & 0.000 & 0.015 & 0.409 & 0.788 & 0.660\\
castle & 0.000 & 0.025 & 0.163 & 0.538 & 0.750 & 0.800\\
chain & 0.000 & 0.154 & 1.731 & 2.385 & 2.654 & 0.260\\
church & 0.000 & 0.029 & 0.353 & 0.912 & 1.088 & 0.680\\
cradle & 0.000 & 0.000 & 0.100 & 0.800 & 1.925 & 0.400\\
electric fan & 0.000 & 0.000 & 0.273 & 0.580 & 0.625 & 0.880\\
go-kart & 0.000 & 0.000 & 0.011 & 0.319 & 0.702 & 0.940\\
holster & 0.000 & 0.013 & 0.066 & 0.395 & 0.658 & 0.760\\
ladle & 0.000 & 0.081 & 0.291 & 0.639 & 0.663 & 0.860\\
lifeboat & 0.000 & 0.117 & 0.596 & 0.766 & 0.702 & 0.940\\
loupe & 0.000 & 0.130 & 0.761 & 1.391 & 1.717 & 0.460\\
paper towel & 0.000 & 0.000 & 0.068 & 0.405 & 0.703 & 0.740\\
ping-pong ball & 0.000 & 0.133 & 0.389 & 0.667 & 0.711 & 0.900\\
punching bag & 0.000 & 0.214 & 0.600 & 0.914 & 1.000 & 0.700\\
saltshaker & 0.000 & 0.106 & 0.576 & 0.742 & 0.788 & 0.660\\
sax & 0.000 & 0.103 & 0.515 & 0.632 & 0.588 & 0.680\\
slot & 0.000 & 0.117 & 0.511 & 0.713 & 0.713 & 0.940\\
spotlight & 0.000 & 0.033 & 0.133 & 1.033 & 1.633 & 0.300\\
teapot & 0.000 & 0.150 & 0.625 & 0.886 & 0.913 & 0.800\\
tram & 0.000 & 0.020 & 0.200 & 1.000 & 1.260 & 0.500\\
table lamp & 0.000 & 0.012 & 0.105 & 0.360 & 0.674 & 0.860\\
vending machine & 0.000 & 0.011 & 0.076 & 0.050 & 0.674 & 0.920\\
wool & 0.000 & 0.000 & 0.075 & 0.525 & 1.350 & 0.400\\
traffic light & 0.000 & 0.057 & 0.341 & 0.670 & 0.841 & 0.880\\
ice lolly & 0.000 & 0.045 & 0.341 & 0.727 & 0.807 & 0.880\\
bagel & 0.000 & 0.029 & 0.286 & 1.000 & 1.214 & 0.700\\
hotdog & 0.000 & 0.000 & 0.012 & 0.267 & 0.616 & 0.860\\
spaghetti squash & 0.000 & 0.000 & 0.012 & 0.159 & 0.671 & 0.820\\
acorn & 0.000 & 0.104 & 0.531 & 0.739 & 0.792 & 0.960\\
rose hip & 0.000 & 0.065 & 0.261 & 0.696 & 0.935 & 0.920\\
coral fungus & 0.000 & 0.000 & 0.010 & 0.188 & 0.521 & 0.960\\
\hline
\textbf{mPW} & 0.000 & 0.056 & 0.322 & 0.701 & 0.936 & 0.761\\
\hline
\textbf{STD} & 0.000 & 0.057 & 0.308 & 0.347 & 0.405 & 0.188\\
\bottomrule
\end{tabular}
}
\end{center}
\end{table*}

\section{Detailed Predictive Power Results for Naturally Trained DNNs}
\label{sec:appd_b}

Here, we show the detailed predictive powers of the class-wise patterns revealed by our method for all 10 CIFAR-10 classes and 50 randomly selected ImageNet classes.
The experimental settings follows Section 4.1 in the main text: ResNet-50 models trained on natural CIFAR-10 and ImageNet. The patterns are searched based on 20\% of the test images, and the predictive power (defined in Section 4 in the main text) is computed on all (e.g. 100\%) test images. In Figure 2 in the main text, we have shown the mean predictive power over all 10 classes of CIFAR-10 and 50 classes of ImageNet, here we show the detailed predictive power for each class in Table \ref{tab:a_1} and Table \ref{tab:a_2}. Note that our method only searches one pattern for each class, and the pattern size is set to be 1\% - 10\% of the image size. Positive sampling of the canvases are used in these experiments.

As shown in Table \ref{tab:a_1} and \ref{tab:a_2}, the predictive power is extremely low when the pattern size is below 5\% of the image size. This confirms our findings in the backdoor trigger recovery experiment in Section 4.3 in the main text: on naturally trained models, small patterns that are of 1\% the image size are not predictive unless the model has learned a backdoor trigger. At pattern size 5\%, the predictive power on CIFAR-10 is 0.59, which means our method has found a pattern that can represent 59\% of the original accuracy on the target class. While these patterns exhibit different predictive power on different classes, the variation is fairly low. This indicates that our method can found representative and predictive patterns reliably for different classes. 
At pattern size 10\%, the mean predictive power across all 10 CIFAR-10 classes is 0.875, that is, the pattern can represent 87.5\% of the original accuracy on average. 
Large patterns are generally more predictive, but contains more background noise. As shown in Table \ref{tab:a_2}, the results on ImageNet are similar to that on CIFAR-10, except that for the same pattern size, the patterns are more predictive on ImageNet and the variation is higher. This is because ImageNet has more diverse images in the same class than CIFAR-10.

\section{Patterns Found with Different Nontarget-class Subsets}
\label{sec:appd_c2}

\begin{figure}[t]
\begin{center}
\includegraphics[width=1\linewidth]{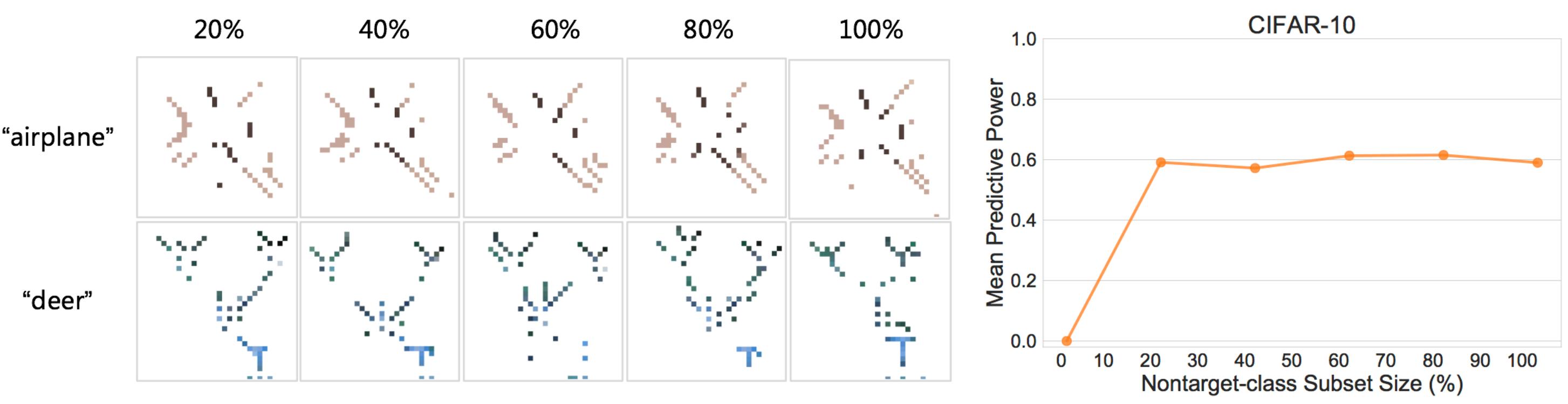}
\end{center}
   \caption{Left grid: Patterns found by our methods under different nontarget-class subset size (\% to the test set), for ``airplane" and ``deer" classes in CIFAR-10. Right plot: The mean predictive power of the class-wise patterns on CIFAR-10. The pattern size is fixed to 5\% of the image size.}
\label{l_fig_16}
\end{figure}

Here, we show the patterns revealed by our method with different sizes of the nontarget-class subset selected from the original test set.
We apply our method to search for the patterns with different nontarget-class subsets of sizes ranging from 20\% to 100\% of the test set. Note that the predictive powers of the patterns found in each setting are still computed on the entire (i.e. 100\%) test set.
This experiment is conducted on a ResNet-50 model trained on natural CIFAR-10 dataset, and the pattern size is fixed to 5\% of the image size.
The class-wise patterns and their predictive powers are illustrated in Figure \ref{l_fig_16}.
As the results show, the patterns are fairly consistent under different subset sizes (left figure), and 20\% of the test set is sufficient for our method to find highly predictive patterns (right figure).

\section{Predictive Powers of Patterns Found on Different Positive Canvases} \label{sec:appd_d3}

Here, we further show how different \emph{positive} canvases affect the predictive power. We test positive canvases selected from the top-$N$ most confident and correctly classified images. We reveal the patterns learned by ResNet-50 models trained on natural (clean) CIFAR-10 and ImageNet datasets. Same as in Section 4.1 in the main text, here we search for the pattern on $\gD_{n}$ and test its predictive power on $\gG_{test}\setminus \gD_{n}$. The results are reported in Table \ref{tab:a_3} and Table \ref{tab:a_4}, where it shows that patterns found on the top-1 positive canvas are not necessary the most predictive ones. As we have shown in Figure 3 in the main text, different sampling strategies can reveal similar shapes, but the details and the locations of the shapes can be quite different. This is also why we suggest to use $N$ (i.e. $N$=5) different canvases to help find the most predictive patterns.

\begin{table*}[t]
\caption{Predictive power of class-wise patterns revealed by different positive canvases on a naturally trained ResNet-50 on CIFAR-10. The canvases are selected using positive sampling based on the top-$N$ ($N$=5) most confident and correctly classified images. Pattern size is fixed to 5\% image size. \textbf{STD}: the standard deviation of the predictive power over all 5 canvases. For each class, the predictive power of the best canvas is highlight in \textbf{bold}.}
\label{tab:a_3}
\begin{center}
\setlength{\tabcolsep}{4mm}{
\begin{tabular}{c|ccccc|c}
\toprule
\multirow{2}{*}{\textbf{Class}} & \multicolumn{5}{c|}{\textbf{Positive Canvas}} & \multirow{2}{*}{\textbf{STD}} \\
 & top-1 & top-2 & top-3 & top-4 & top-5 &\\
\midrule
airplane & 0.473 & 0.651 & \textbf{0.745} & 0.729 & 0.392 & 0.141\\
bird & \textbf{0.665} & 0.623 & 0.470 & 0.220 & 0.591 & 0.161\\
dear & 0.622 & \textbf{0.698} & 0.659 & 0.511 & 0.553 & 0.068\\
frog & 0.135 & \textbf{0.383} & 0.377 & 0.246 & 0.276 & 0.092\\
Ship & 0.312 & 0.262 & 0.507 & 0.550 & \textbf{0.589} & 0.132\\
\bottomrule
\end{tabular}
}
\end{center}
\end{table*}

\begin{table*}[t]
\caption{Predictive power of class-wise patterns revealed by different positive canvases on a naturally trained ResNet-50 on ImageNet. The canvases are selected using positive sampling based on the top-$N$ ($N$=5) most confident and correctly classified images. Pattern size is fixed to 5\% image size. \textbf{STD}: the standard deviation of the predictive power over all 5 canvases. For each class, the predictive power of the best canvas is highlight in \textbf{bold}.}
\label{tab:a_4}
\begin{center}
\setlength{\tabcolsep}{4mm}{
\begin{tabular}{c|ccccc|c}
\toprule
\multirow{2}{*}{\textbf{Class}} & \multicolumn{5}{c|}{\textbf{Positive Canvas}} & \multirow{2}{*}{\textbf{STD}} \\
 & top-1 & top-2 & top-3 & top-4 & top-5 & \\
\midrule
tench & 0.599 & \textbf{0.609} & 0.433 & 0.420 & 0.476 & 0.081\\
tibetan terrier & 0.760 & 0.793 & 0.707 & \textbf{0.861} & 0.682 & 0.064\\
academic gown & 0.411 & 0.572 & 0.301 & \textbf{0.632} & 0.596 & 0.126\\
hook & \textbf{1.271} & 0.823 & 0.511 & 0.591 & 0.615 & 0.274\\
Slot & \textbf{0.713} & 0.702 & 0.590 & 0.655 & 0.652 & 0.044\\
\bottomrule
\end{tabular}
}
\end{center}
\end{table*}

\section{Predictive Powers of Patterns Found by Different Searching Methods}
\label{sec:appd_e}

Table \ref{tab:a_5} and Table \ref{tab:a_6} report the the mean predictive power of the patterns revealed by the 4 methods: 1) our original method (Equation 2 in the main text), 2) UAP, 3) the TV variant of our method and 4) the learned canvas variant of our method. 

In Table \ref{tab:a_5}, the mean predictive power is computed over all 10 CIFAR-10 classes or the 50 randomly selected ImageNet classes. For UAP, we use the attack success rate (ASR) as the predictive power. As can be inferred from Table \ref{tab:a_5}, the predictive powers of our method are similar to UAP with maximum perturbation $\epsilon$=8. UAP has much higher predictive power at $\epsilon$=16 or $\epsilon$=32, as expected.
For the TV version of our method, increasing the regularization strength from $\beta$=0.2 to $\beta$=1 tends to find less predictive patterns. This is because TV removes the predictive background noise. This confirms that background noise also plays an important role in DNNs \citep{DBLP:conf/nips/IlyasSTETM19}, which also aligns with our findings in the adversarial setting in Section 4.4 in the main text.

Table \ref{tab:a_6} reports the mean predictive powers of the patterns found on sampled (by our original method) versus learned canvases. The predictive power is tested on the entire test set of either CIFAR-10 or ImageNet. Overall, the predictive powers of the patterns found on learned canvases are similar to those found on fixed canvases, though there are certain variations. Surprisingly, the learned (perturbed) canvas does not lead to more predictive patterns, considering the high effectiveness of adversarial perturbations. Upon further investigation, we find that this is due to three reasons: 1) the pattern revealed when using the learned canvas is a class-wise universal adversarial pattern, which is less effective (predictive) than a sample-wise adversarial perturbation; 2) only 5\% of the perturbed canvas is used to extract the pattern, which significantly reduces the adversarial (predictive) effect; 3) the pattern needs to be attached to other clean images to compute the predictive power.  This is a transfer scenario that also reduces the adversarial effect. This eventually results in similarly predictive patterns for both the fixed canvas and the learned canvas.

\begin{table*}[t]
\caption{Attack success rate and predictive power of class-wise patterns generated by universal adversarial perturbation (UAP) \citep{moosavi2017universal} and our method. For our method, we also test the use of the total-variation (TV) \citep{fong2017interpretable} regularization along with our $L_1$ regularization (this variant of our method is defined in Equation 3 in the main text). The UAP pattern size is 100\% of the image size while our patterns ($L_1$ or TV) are 5\% of the image size.}
\label{tab:a_5}
\begin{center}
\setlength{\tabcolsep}{1.2mm}{
\begin{tabular}{c|ccc|ccc}
\toprule
Methods & UAP ($\epsilon$=8) & UAP ($\epsilon$=16) & UAP ($\epsilon$=32) & Ours & Ours+TV ($\beta$=0.2) & Ours+TV ($\beta$=1)\\
\midrule
CIFAR-10 & 0.621 & 0.967 & 0.982 & 0.609 & 0.580 & 0.302 \\
ImageNet & 0.607 & 0.683 & 0.944 & 0.623 & 0.597 & 0.272 \\
\bottomrule
\end{tabular}
}
\end{center}
\end{table*}

\section{Comparison to Grad-CAM Attention Maps}
\label{sec:appd_g}
\begin{figure}[t]
\centering
\includegraphics[width=0.95\linewidth]{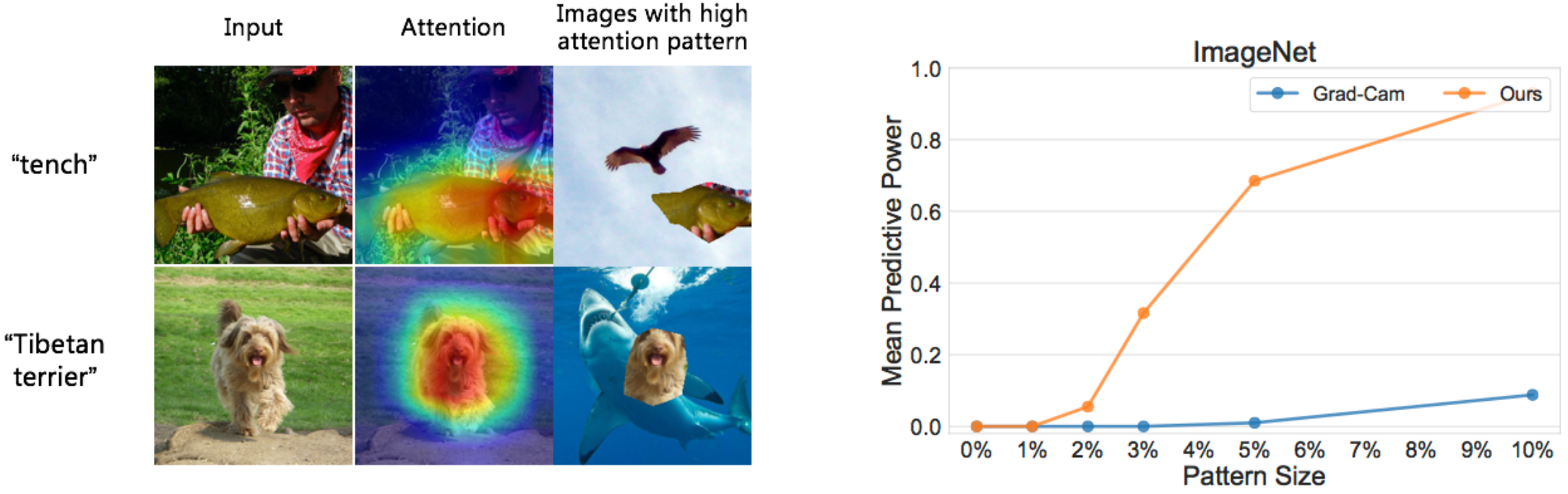}
\vspace{-0.1in}
   \caption{\emph{Left grid}: input image (first column), attention map (second column) and images (third column) attached with high attention patterns extracted from the input image. \emph{Right plot}: the mean predictive power of high attention patterns over 50 ImageNet target classes.
   The attention pattern size is 10\% of the image size.}
\label{l_fig_11}
\vspace{-0.1in}
\end{figure}

\noindent\textbf{Predictive Power Comparison.}
Here, we compare the predictive power of the class-wise patterns found by our method to the key areas identified by using the attention map. This is somewhat not a fair comparison since attention maps were designed for sample-wise explanations.
However, we believe it is interesting to see if high attention areas are also predictive of the class. 
We use Grad-CAM \citep{DBLP:journals/corr/SelvarajuDVCPB16} to extract the high attention area of the positively sampled canvas image for each of the 50 ImageNet target classes (class names are in Table \ref{tab:a_2}). The size of the high attention area is set to be 10\% of the image.
We then attach the extracted attention area to all non-target class test images to compute its predictive power (the same testing scheme as for our method). 
Figure \ref{l_fig_11} illustrates a few example attention areas (left grid) and the mean predictive power over the 50 classes (right plot).  We find that the predictive power of the attention patterns are considerably low. This indicates that sample-wise attention maps may not be a good representation of the class-wise knowledge learned by the model. 
As the attention visualization shown in Appendix \ref{sec:appd_g}, the patterns identified by our method indeed have a noticeable correlation to the class and can consistently attract the network's attention when attached to different images.

\begin{figure}[t]
\begin{center}
\includegraphics[width=1\linewidth]{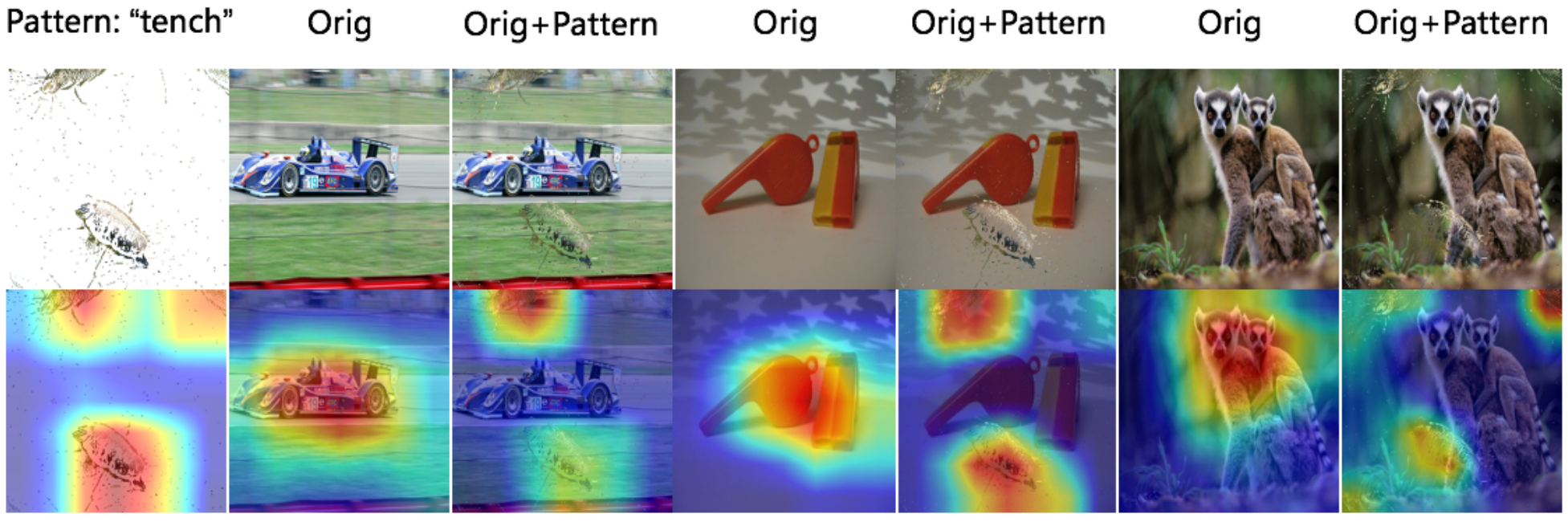}
\end{center}
   \caption{The attention (visualized by Grad-CAM) shift of a ResNet-50 model (on ImageNet) when our class-wise pattern is attached to different images. The pattern is from ``n01440764"(``tench")}
\label{l_fig_10}
\end{figure}

\noindent\textbf{Attention Shift When a Class-wise Pattern is Attached.}
We use the attention map to check the network's attention shift when the class-wise pattern found by out method is attached to an image.
As shown in Figure \ref{l_fig_10}, we find that class-wise pattern has very strong attention around the areas that contain clear shapes, and the attention of the network is significantly shifted towards those areas when the pattern is attached to different images.
This confirms that the patterns identified by our method are indeed universal patterns that carry consistent information.

\begin{table*}[t]
\caption{Predictive power of class-wise patterns revealed on sampled (fixed) vs learned canvases (by Equation 4 in the main text) for naturally trained ResNet-50s on CIFAR-10 and ImageNet. Pattern size is fixed to 5\% image size.}
\label{tab:a_6}
\begin{center}
\setlength{\tabcolsep}{4mm}{
\begin{tabular}{c|ccc|ccc}
\toprule
\multirow{2}{*}{\textbf{Method}} & \multicolumn{3}{c}{\textbf{CIFAR-10 Class}} & \multicolumn{3}{c}{\textbf{ImageNet Class}} \\
 & airplane & deer & frog & tabby & analog clock & bagel \\
\midrule
Sampled Canvas & 0.745 & 0.698 & 0.383 & 0.547 & 0.711 & 1.000 \\
Learned Canvas & 0.760 & 0.659 & 0.515 & 0.164 & 0.667 & 0.816\\
\bottomrule
\end{tabular}
}
\end{center}
\end{table*}

\end{document}

%% file: math_commands.tex
\usepackage{amsmath,amsfonts,bm}

\def\eqref#1{equation~(\ref{#1})}
\def\Eqref#1{Equation~(\ref{#1})}

\def\1{\bm{1}}

\def\vm{{\bm{m}}}

\def\vp{{\bm{p}}}

\def\vx{{\bm{x}}}

\DeclareMathAlphabet{\mathsfit}{\encodingdefault}{\sfdefault}{m}{sl}
\SetMathAlphabet{\mathsfit}{bold}{\encodingdefault}{\sfdefault}{bx}{n}

\def\gD{{\mathcal{D}}}

\def\gG{{\mathcal{G}}}

\def\gL{{\mathcal{L}}}

\def\gP{{\mathcal{P}}}

